\documentclass[letterpaper,10pt,conference]{ieeeconf}

\IEEEoverridecommandlockouts
\usepackage{amsmath,amssymb}
\usepackage{graphicx}
\usepackage{booktabs}
\usepackage{siunitx}
\usepackage{algorithm}
\usepackage{algpseudocode}
\usepackage{float}
\usepackage{cite}
\usepackage[hyphens]{url}
\usepackage[capitalize]{cleveref}

\title{\LARGE\bfseries
Memory-Guided Trust-Region Bayesian Optimization (MG-TuRBO) for High Dimensions}

\author{%
Abhilasha Saroj, Shaked Regev, Guanhao Xu, Jinghui Yuan, Xiangyong Luo, and Chieh (Ross) Wang%
\thanks{This manuscript has been authored by UT-Battelle, LLC, under contract DE-AC05-00OR22725 with the US Department of Energy (DOE). The US government retains and the publisher, by accepting the article for publication, acknowledges that the US government retains a nonexclusive, paid-up, irrevocable, worldwide license to publish or reproduce the published form of this manuscript, or allow others to do so, for US government purposes. DOE will provide public access to these results of federally sponsored research in accordance with the DOE Public Access Plan ( https://www.energy.gov/doe-public-access-plan )}
}

\begin{document}
\maketitle
\thispagestyle{empty}
\pagestyle{empty}

\begin{abstract}
Traffic simulation and digital-twin calibration is a challenging optimization problem with a limited simulation budget. Each trial requires an expensive simulation run, and the relationship between calibration inputs and model error is often nonconvex, and noisy. The problem becomes more difficult as the number of calibration parameters increases. We compare a commonly used automatic calibration method, a genetic algorithm (\textbf{GA}), with Bayesian optimization methods (\textbf{BOM}s): classical Bayesian optimization (\textbf{BO}), Trust-Region BO (\textbf{TuRBO}), \textbf{Multi-TuRBO}, and a proposed Memory-Guided TuRBO (\textbf{MG-TuRBO}) method. We compare performance on 2 real-world traffic simulation calibration problems with 14 and 84 decision variables, representing lower- and higher-dimensional (\textbf{14D and 84D}) settings. For BOMs, we study two acquisition strategies, Thompson sampling and a novel adaptive strategy. We evaluate performance using final calibration quality, convergence behavior, and consistency across runs. The results show that BOMs reach good calibration targets much faster than GA in the lower-D problem. MG-TuRBO performs comparably in our 14D setting, it demonstrates noticeable advantages in the 84D problem, particularly when paired with our adaptive strategy. Our results suggest that MG-TuRBO is especially useful for high-D traffic simulation calibration and potentially for high-D problems in general.
\end{abstract}

\section{Introduction}
Traffic simulation models are becoming increasingly important for traffic operation, planning, and safety analysis as sensing, connected infrastructure, and data pipelines continue to improve\cite{saroj-DT}. Their value depends on how well they represent reality, i.e., how well they are calibrated to observed traffic conditions. If key inputs such as unknown traffic inflows, turn ratios, or behavior-related parameters are not properly calibrated, the digital twin can produce biased network states and unreliable decisions. Here, we study traffic simulation calibration as an expensive black-box optimization problem. 
    
    

Traffic microsimulation calibration is difficult because the objective is stochastic, nonconvex, and costly to evaluate. 
In practice, calibration is typically assessed by agreement with observed counts, speeds using fitness measures 
\cite{FHWA_TAT_Vol4_Calibration}. 
Prior studies and practitioner guidance have established calibration workflows for common simulation platforms, while also showing that calibration becomes harder as network size, congestion, and parameter dimension increase \cite{SUMO2018}. Limited simulation budgets make search efficiency a key challenge.

Metaheuristic methods remain widely used for traffic simulation calibration because they are flexible, robust, and practical when gradients are unavailable \cite{Karimi2019,WSC_comparison}.
Among them, Genetic Algorithm (\textbf{GA}) is especially common and has shown practical value across many calibration settings \cite{Karimi2019}. However, these methods often require many simulator evaluations because they do not explicitly model the response surface. This has motivated interest in more sample-efficient methods for expensive calibration problems. Bayesian optimization methods (\textbf{BOM}s) are attractive in this setting because they use a Gaussian process (\textbf{GP}) with an acquisition function to guide evaluations toward promising regions of the search space \cite{Srinivas2010GPUCB}.
BOMs have also shown promise in traffic simulation calibration \cite{osorio2017scalable}. 


However, standard global Bayesian Optimization (\textbf{BO}) often becomes less effective as dimension increases because surrogate modeling and acquisition optimization become harder over large search spaces \cite{TAY2022210}. Trust-Region BO (\textbf{TuRBO}) improves scalability by restricting search to local trust regions, and \textbf{Multi-TuRBO} extends this idea by running multiple trust regions to improve diversity \cite{Eriksson2019TuRBO}. These methods use Thompson sampling (posterior-sample-based candidate selection) acquisition strategy \cite{pmlr-v84-kandasamy18a}. They are well suited to traffic simulation calibration, where objectives are noisy and multimodal and evaluation budgets are limited. Still, after trust-region collapse, restart decisions can return the search to regions with similar local minima, reducing efficiency in higher-D settings.

We introduce \textbf{Memory-Guided TuRBO (MG-TuRBO)}
and an \textbf{adaptive acquisition strategy}, which extend TuRBO in complementary ways. MG-TuRBO uses evaluation history to cluster sampled points in normalized design space into candidate basins, computes basin-level quality and visitation statistics from observed objective values, and uses these statistics to select restart centers from promising but under-explored basins while filtering clearly weak basins. This basin-aware restart policy reduces redundant rediscovery of similar local optima after trust-region collapse. Our adaptive strategy is a time-varying weighted combination of improvement and predictive uncertainty. It provides explicit control of the exploration--exploitation trade-off during search.

\section{TRAFFIC SIMULATION CALIBRATION AS A BLACK-BOX OPTIMIZATION PROBLEM}

\subsection{Problem Formulation}

We formulate traffic simulation calibration as an expensive black-box optimization problem. Calibration quality is evaluated by comparing simulated traffic counts with observed counts using the Geoffrey E. Havers (GEH) statistic, which is a widely used traffic-model calibration measure 
\cite{fhwa_sr99_2021}.
Let $o_n$ denote the observed count and $s_n(\mathbf{x};\omega)$ denote the simulated count for candidate parameter vector $\mathbf{x}$ under simulator realization $\omega$, such as a given random seed. For each target, GEH is computed as
\begin{equation}
\mathrm{GEH}_n(\mathbf{x};\omega)=
\sqrt{
\frac{2\left(s_n(\mathbf{x};\omega)-o_n\right)^2}
{s_n(\mathbf{x};\omega)+o_n}
}.
\label{eq:geh}
\end{equation}

We use mean GEH,
$\bar{g}(\mathbf{x};\omega)\doteq\frac{1}{N}\sum_{n=1}^{N}\mathrm{GEH}_n(\mathbf{x};\omega)$, where $N$ is the total number of count comparisons,
as the primary network-level calibration metric. Lower values indicate better agreement of simulated and observed counts. We also track the fraction of count targets with GEH $\leq 5$ as a commonly used calibration-compliance indicator \cite{fhwa_sr99_2021}.

The decision vector is $\mathbf{x}\in\mathcal{X}\subset\mathbb{R}^{d}$
In this study, the calibration dimension $d$ is either 14 or 84. Each parameter satisfies box constraints $\ell_j \le x_j \le u_j$ 
For a given optimization run, the calibration problem is
\begin{equation}
\min_{\mathbf{x}\in\mathcal{X}} \ \bar{g}(\mathbf{x};\omega) \doteq \frac{1}{N}\sum_{n=1}^{N}\mathrm{GEH}_n(\mathbf{x};\omega).
\label{eq:optimization_problem}
\end{equation}
Here, we evaluate each candidate solution using a single simulator realization, $\omega$. The optimizer minimizes the corresponding mean GEH rather than an expectation across multiple seeds. The total number of simulator evaluations is limited to a finite budget $B$, and we apply this same budget to all compared algorithms, including GA, for a fair comparison. In the reported 14D and 84D experiments, all methods optimize $\bar{g}(\mathbf{x};\omega)$, while GEH compliance is tracked for analysis and target checking rather than enforced as a hard constraint. BO, TuRBO, Multi-TuRBO, and MG-TuRBO use \cref{eq:optimization_problem} within a surrogate-guided search framework. GA uses the same scalar objective and budget without surrogate modeling or an acquisition function.

\subsection{Benchmark Networks and Calibration Setup}
\begin{figure}[htbp]
    \centering
    \includegraphics[width=0.95\linewidth]{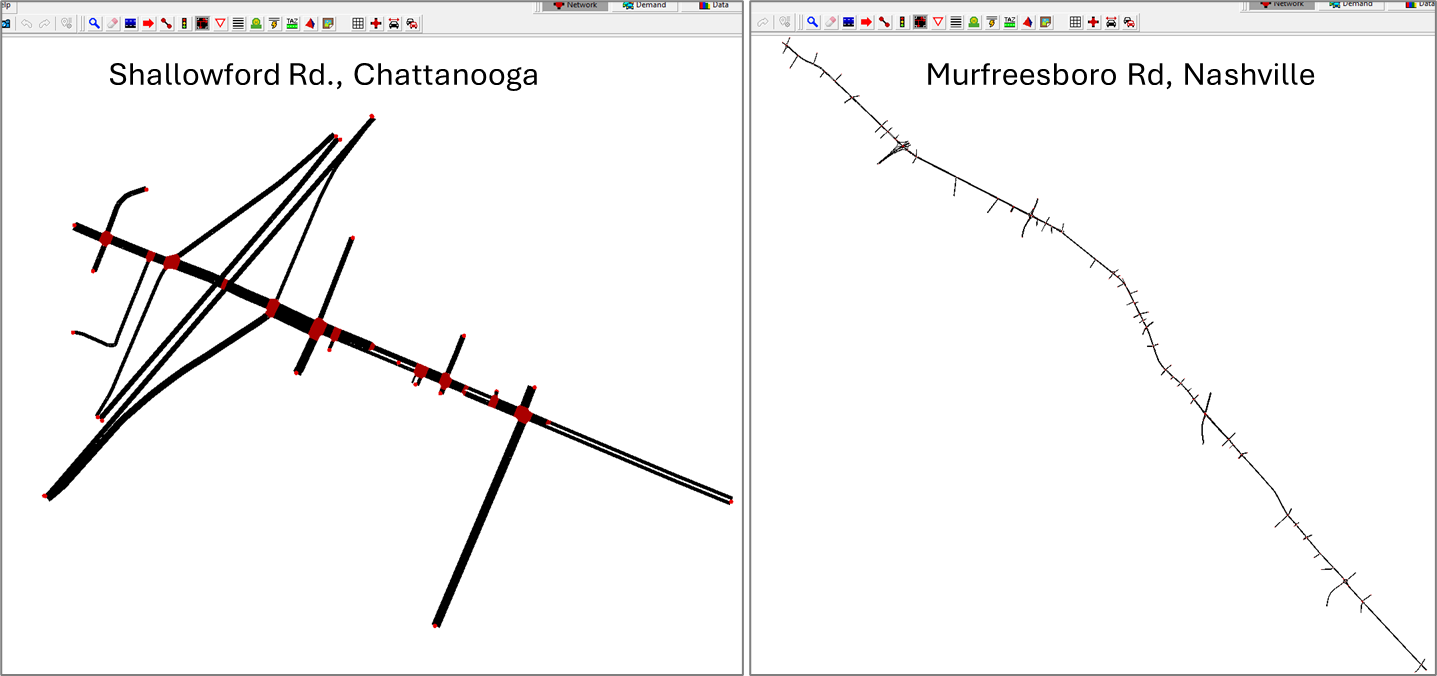}
    \caption{SUMO models of the calibration networks: (Left) 14D Shallowford Road network (Right) 84D Murfreesboro Road corridor network.}
    \label{fig:study_networks}
\end{figure} 
We evaluate the calibration methods on two traffic digital twin networks developed in the Real-Twin platform \cite{xu2025automated} and implemented in \textit{Simulation of Urban MObility} (SUMO) \cite{Krajzewicz2012SUMO}, as shown in \cref{fig:study_networks}. The first network models the Shallowford Road corridor in Chattanooga, Tennessee, and defines a 14D calibration problem. 
The second network models the Murfreesboro Pike corridor in Nashville, Tennessee, and defines an 84D calibration problem. 
These 2 networks allow us to compare optimizer performance across moderate- and high-D calibration settings.





\section{OPTIMIZATION METHODS AND EXPERIMENT DESIGN}

All methods minimize \cref{eq:optimization_problem} under a fixed budget of $B$ simulator evaluations. Within each network, all methods use the same parameter bounds and initial design to ensure fair comparison. We compare a baseline metaheuristic method GA with BOMs: standard BO, TuRBO, Multi-TuRBO, and MG-TuRBO. 

\subsection{Bayesian Optimization}
BO is a sequential and sample-efficient method for expensive black-box optimization. It fits a probabilistic surrogate to the objective function and uses that surrogate to select the next simulator evaluation. We use a Gaussian Process (\textbf{GP)}:
\begin{equation}
f(\mathbf{x}) \sim \mathcal{GP}(\mu(\mathbf{x}), k(\mathbf{x},\mathbf{x}')),
\label{eq:gp}
\end{equation}
where $\mu(\mathbf{x})$ is the mean function and $k(\mathbf{x},\mathbf{x}')$ is the covariance kernel. Given the observed data $\mathcal{D}_t=\{(\mathbf{x}_i,y_i)\}_{i=1}^{t}$, the GP provides a posterior mean $\mu_t(\mathbf{x})$ and standard deviation $\sigma_t(\mathbf{x})$ for any candidate point $\mathbf{x}$. BO selects the next evaluation using an acquisition strategy:
\begin{equation}
\mathbf{x}_{t+1}=\arg\max_{\mathbf{x}\in\mathcal{X}} \alpha(\mathbf{x}),
\label{eq:bo_next}
\end{equation}
where $\alpha(\mathbf{x})$ is computed from the GP posterior and favors points that appear promising or remain uncertain and therefore informative to evaluate.

At each iteration, we approximate \cref{eq:bo_next} using a finite Sobol candidate set \cite{scipy} in the normalized search space, evaluate the acquisition strategy on that set, and select the highest-scoring candidate for evaluation. Before GP fitting, calibration parameters and objective values are standardized to mean $0$ and variance $1$ to improve numerical stability.




\subsection{TuRBO}
TuRBO addresses high-dimensional optimization by restricting the BO search in \cref{eq:bo_next} to a local trust region rather than the full domain \cite{Eriksson2019TuRBO}. Instead of evaluating the acquisition strategy over candidates drawn from the entire normalized space, TuRBO generates candidates within a hyperrectangular trust region $\mathcal{T}_t \subseteq [0,1]^d$ centered at the current best observation. \cref{eq:bo_next} is applied locally within $\mathcal{T}_t$ rather than globally over $\mathcal{X}$. The trust region is controlled by a size parameter $\delta_t \in [\delta_{\min}, \delta_{\max}]$ that adapts to local progress. After $\tau_{\text{succ}}$ consecutive improvements, the region expands $\delta_{t+1} \leftarrow \min(\delta_{\max}, 2\delta_t)$. After $\tau_{\text{fail}}$ consecutive non-improving evaluations, it contracts $\delta_{t+1} \leftarrow \max(\delta_{\min}, \delta_t/2)$.

If the region shrinks to $\delta_{\min}$ without improvement, TuRBO restarts from a new random location and resets $\delta$. This helps the method escape poor local basins while concentrating evaluations in promising neighborhoods. By restricting candidate generation to a local region, TuRBO reduces the effective search space seen by the GP and improves search efficiency in higher-dimensional settings.

\subsection{Multi-TuRBO}
Multi-TuRBO extends TuRBO by maintaining $M$ independent trust regions in parallel \cite{Eriksson2019TuRBO}. Each region $\mathcal{T}_t^{(m)}$ has its own center (the best point found within that region), size parameter $\delta_t^{(m)}$, and success/failure counters. At each iteration, one candidate is proposed from each active region:
\begin{equation}
\mathbf{x}_{t+1}^{(m)}=\arg\max_{\mathbf{x}\in \mathcal{T}_t^{(m)}} \alpha(\mathbf{x}), \qquad m=1,\dots,M.
\label{eq:multiturbo_acq}
\end{equation}
All $M$ candidates are evaluated in the same iteration (using parallel simulator instances when available), and the data are pooled into a single dataset $\mathcal{D}_t$ for GP training.

Each region evolves independently according to the expansion, contraction, and restart rules described for TuRBO. When a region collapses to $\delta_{\min}^{(m)}$ and triggers a restart, it is re-initialized at a random location while the other regions continue their local search. This parallel multi-region design increases exploration diversity and reduces the risk of premature convergence to a single local basin. By distributing the evaluation budget across $M$ simultaneous local searches, Multi-TuRBO can discover multiple promising regions and adaptively allocate effort based on each region's progress.

\subsection{Memory-Guided TuRBO}

MG-TuRBO extends Multi-TuRBO by modifying only the restart step. During normal search, each trust region follows the same local acquisition-based update and trust-region adaptation rules as Multi-TuRBO. The difference appears when a trust region collapses to its minimum size. Instead of restarting from a random location, MG-TuRBO uses historical evaluations to identify promising basins and restart in regions that are both high quality and under-explored. \cref{alg:mgturbo} summarizes the procedure.

Let $\mathcal{D}_t=\{(\mathbf{x}_i,y_i)\}_{i=1}^{t}$ denote the set of evaluated points up to iteration $t$, where $y_i=f(\mathbf{x}_i)$. Every $\tau_{\text{recluster}}$ evaluations, MG-TuRBO clusters the normalized design points in $\mathcal{D}_t$ into $K$ basins, denoted by $\{\mathcal{B}_1,\dots,\mathcal{B}_K\}$. For each basin $\mathcal{B}_k$, it computes the basin quality
\begin{equation}
q_k=\min_{\mathbf{x}\in\mathcal{B}_k} f(\mathbf{x}),
\label{eq:basin_quality}
\end{equation}
which is the best objective value found in that basin, and the basin population $n_k = |\mathcal{B}_k|$, which counts how many evaluated points currently belong to basin $k$.

When a trust region collapses, MG-TuRBO first removes clearly poor basins. It keeps only the basins that satisfy
\begin{equation}
\mathcal{K}_{\text{keep}}=\left\{k: q_k \le q_{\text{best}}+\gamma \sigma_q\right\},
\label{eq:basin_filter}
\end{equation}
where $q_{\text{best}}=\min_k q_k$, $\sigma_q$ is the standard deviation of basin qualities. $\gamma$ controls how far from the current best a basin may be before it is discarded. MG-TuRBO then scores the retained basins using
\begin{equation}
s_k = w_E \cdot \log\left(1 + \frac{1}{n_k + 1}\right) + w_P \cdot \exp\left(-\frac{q_k - q_{\text{best}}}{\sigma_q}\right).
\label{eq:basin_score}
\end{equation}
The first term favors basins with fewer assigned samples and therefore promotes exploration. The second term favors basins whose best value is close to the current global best and therefore promotes exploitation. The weights $w_E$ and $w_P$ control this balance. In this study, we use $w_E=0.3$ and $w_P=0.7$, which places greater emphasis on exploitation.

After scoring, MG-TuRBO selects the restart basin as
\begin{equation}
k^*=\arg\max_{k\in\mathcal{K}_{\text{keep}}} s_k,
\label{eq:best_basin}
\end{equation}
and restarts the collapsed trust region from the best previously evaluated point in that basin:
\begin{equation}
\mathbf{x}_{\text{restart}}=\arg\min_{\mathbf{x}\in\mathcal{B}_{k^*}} f(\mathbf{x}).
\label{eq:restart_point}
\end{equation}
Thus, \cref{eq:basin_filter,eq:restart_point} define the memory-guided restart logic in \cref{alg:mgturbo}.

Compared with TuRBO and Multi-TuRBO, MG-TuRBO adds three memory-based components: periodic basin discovery, quality-aware basin filtering, and population-aware restart selection. 
Multi-TuRBO improves diversity through multiple trust regions, but still restarts randomly. MG-TuRBO instead uses accumulated search history to guide restart toward promising regions that have not yet been heavily sampled. This design aims to reduce repeated rediscovery of similar local optima and improve budget efficiency in multimodal high-dimensional calibration problems.

\begin{algorithm}[t]
\caption{MG-TuRBO}
\label{alg:mgturbo}
\footnotesize
\begin{algorithmic}[1]
\Require Budget $B$, initial data $\mathcal{D}_0$, number of regions $M$
\Require Trust-region parameters $\delta_{\text{init}}, \delta_{\min}, \delta_{\max}, \tau_{\text{succ}}, \tau_{\text{fail}}$
\Require Basin parameters $K$, $\tau_{\text{recluster}}$, $\gamma$, $w_E$, $w_P$
\State Initialize $M$ trust regions as in Multi-TuRBO
\State $\mathcal{D} \gets \mathcal{D}_0$, \ $t_{\text{cluster}} \gets 0$
\While{$|\mathcal{D}| < B$}
    \If{$|\mathcal{D}| - t_{\text{cluster}} \ge \tau_{\text{recluster}}$}
        \State Cluster $\mathcal{D}$ into basins $\{\mathcal{B}_k\}_{k=1}^K$
        \For{$k = 1,\dots,K$}
            \State $q_k \gets \min_{\mathbf{x}\in\mathcal{B}_k} f(\mathbf{x})$ \Comment{\cref{eq:basin_quality}}
            \State $n_k \gets |\mathcal{B}_k|$
        \EndFor
        \State $t_{\text{cluster}} \gets |\mathcal{D}|$
    \EndIf
    \State Fit GP surrogate on $\mathcal{D}$
    \For{$m = 1,\dots,M$}
        \State Select and evaluate $\mathbf{x}_{\text{next}}^{(m)}$ in $\mathcal{T}^{(m)}$
        \State Add $(\mathbf{x}_{\text{next}}^{(m)}, y_{\text{next}}^{(m)})$ to $\mathcal{D}$
        \State Update center, counters, and $\delta^{(m)}$ as in Multi-TuRBO
        \If{$\delta^{(m)} = \delta_{\min}$}
            \State $q_{\text{best}} \gets \min_k q_k$, \ $\sigma_q \gets \mathrm{std}(\{q_k\})$
            \State $\mathcal{K}_{\text{keep}} \gets \{k : q_k \le q_{\text{best}} + \gamma \sigma_q\}$ \Comment{\cref{eq:basin_filter}}
            \If{$\mathcal{K}_{\text{keep}} = \emptyset$}
                \State Restart region $m$ from current global best
            \Else
                \For{$k \in \mathcal{K}_{\text{keep}}$}
                    \State $s_k \gets w_E \log\!\left(1+\frac{1}{n_k+1}\right) + w_P \exp\!\left(-\frac{q_k-q_{\text{best}}}{\sigma_q}\right)$ \Comment{\cref{eq:basin_score}}
                \EndFor
                \State $k^* \gets \arg\max_{k\in\mathcal{K}_{\text{keep}}} s_k$ \Comment{\cref{eq:best_basin}}
                \State $\mathbf{x}_{\text{restart}} \gets \arg\min_{\mathbf{x}\in\mathcal{B}_{k^*}} f(\mathbf{x})$ \Comment{\cref{eq:restart_point}}
                \State Restart region $m$ at $\mathbf{x}_{\text{restart}}$
            \EndIf
            \State Reset $\delta^{(m)} \gets \delta_{\text{init}}$, $n_{\text{succ}}^{(m)} \gets 0$, $n_{\text{fail}}^{(m)} \gets 0$
        \EndIf
    \EndFor
\EndWhile
\State \Return $\arg\min_{\mathbf{x}\in\mathcal{D}} f(\mathbf{x})$
\end{algorithmic}
\end{algorithm}

\subsection{Acquisition Strategies}
\label{sec:acquisition}

The BOMs use two acquisition strategies in this study: adaptive and Thompson sampling.

\textbf{Adaptive strategy:} This strategy combines expected improvement (\textbf{EI}) with predictive uncertainty~\cite{Regev2025}. Let $f^*=\min_{i=1,\dots,t} y_i$ denote the best objective value observed so far. EI at candidate point $\mathbf{x}$ is
\begin{align}
&  \alpha_{\text{EI}}(\mathbf{x})
=(f^*-\mu_t(\mathbf{x}))\Phi(Z)
+\sigma_t(\mathbf{x})\phi(Z),\label{eq:ei}
\end{align}
with $Z=\frac{f^*-\mu_t(\mathbf{x})}{\sigma_t(\mathbf{x})}$.
$\mu_t(\mathbf{x})$ and $\sigma_t(\mathbf{x})$ are the GP posterior mean and standard deviation.$\Phi(\cdot)$ and $\phi(\cdot)$ are the standard normal cumulative distribution and probability density functions. EI favors points that have a low predicted objective value or high predictive uncertainty.

The adaptive acquisition strategy combines normalized EI and normalized uncertainty $\alpha(\mathbf{x})=(1-\beta_t)\,\widetilde{\mathrm{EI}}(\mathbf{x})+\beta_t\,\widetilde{\sigma}_t(\mathbf{x})$. $\widetilde{\mathrm{EI}}(\mathbf{x})$ and $\widetilde{\sigma}_t(\mathbf{x})$ are normalized versions of EI and predictive uncertainty over the candidate set. The parameter $\beta_t$ controls the exploration--exploitation trade-off. Here, $\beta_t$ follows a time-varying schedule that starts with greater emphasis on exploration and gradually shifts toward exploitation.

\textbf{Thompson sampling:} Thompson sampling \cite{pmlr-v84-kandasamy18a} draws a sample function from the GP posterior over the candidate set and selects the candidate with the best sampled value $\mathbf{x}_{t+1}=\arg\min_{\mathbf{x}\in\mathcal{C}_t} \mathcal{GP}(\mu_t,\sigma_t^2)$, where $\mathcal{C}_t$ denotes the Sobol candidate set at iteration $t$. This naturally balances exploration and exploitation through posterior uncertainty without requiring a weighting parameter.

\subsection{Experiment Design Overview}


For the 14D Chattanooga problem, the total budget is $B=100$ evaluations. All methods also start from the same Sobol initial 20 runs for surrogate model initialization. We run each algorithm 10 times with different random seeds to examine consistency across runs. For the 84D Nashville problem, the total budget is $B=1500$ evaluations with initial 200 runs for for surrogate model initialization.  Due to the higher computational cost, we run each method once.


For TuRBO methods, we use $\delta_{\text{init}}=0.8$, $\delta_{\min}=2^{-5}$, $\delta_{\max}=1.6$, $\tau_{\text{succ}}=2$, and $\tau_{\text{fail}}=5$. Multi-TuRBO uses $M=3$ trust regions. BO, TuRBO, Multi-TuRBO, and MG-TuRBO are each tested with adaptive and Thompson sampling acquisition strategy to study the effects of MG-TuRBO and adaptive acquisition.
\section{RESULTS}
\subsection{14 Dimensional Calibration Optimization}

\begin{figure}[!t]
\centering
\includegraphics[width=\columnwidth]{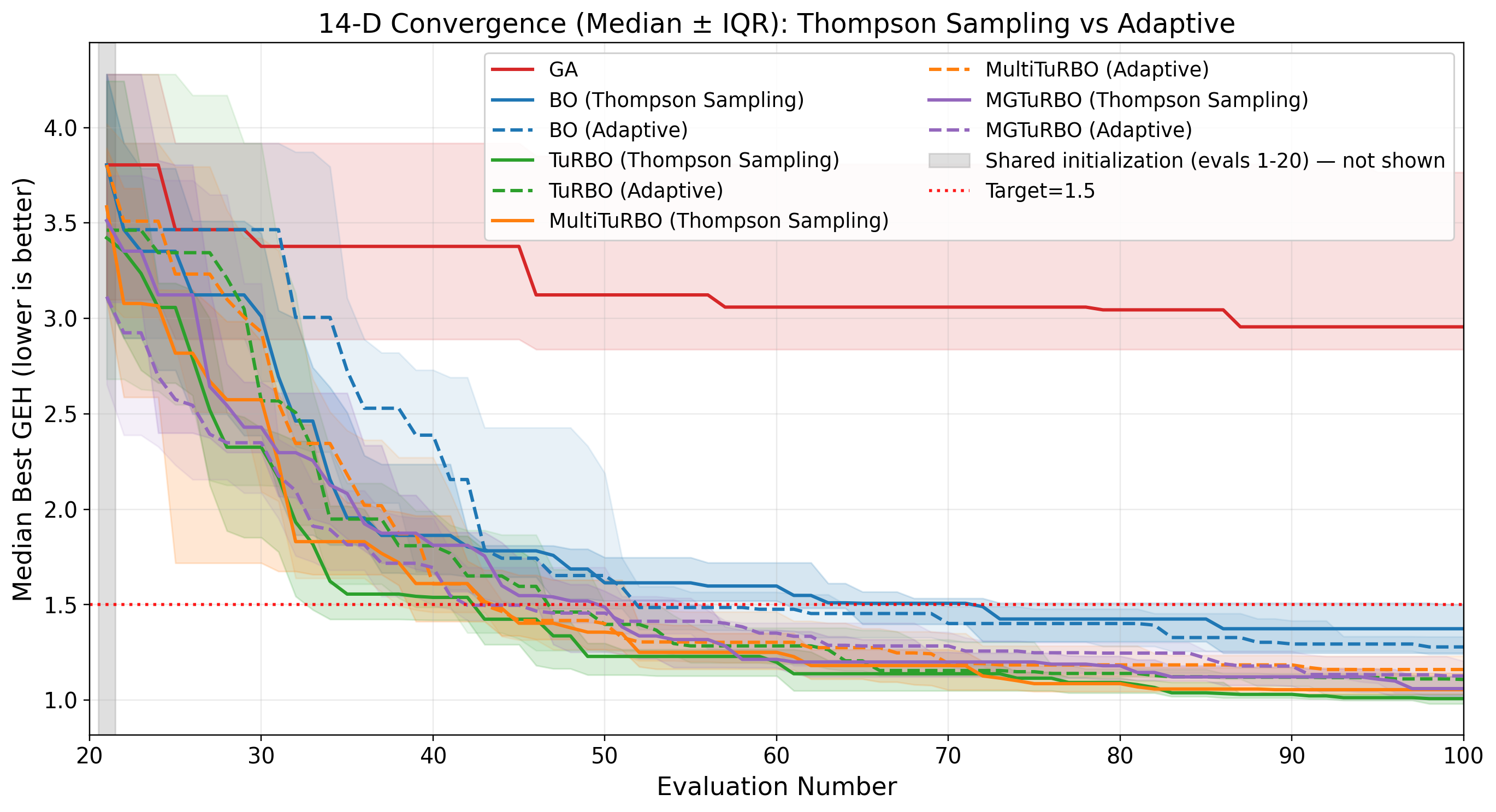}
\caption{Optimization-phase convergence for the 14D network.}
\label{fig:convergence_14d}
\end{figure}

\cref{fig:convergence_14d} shows optimization-phase convergence on the 14D network, reported as median best-observed GEH with interquartile range (IQR) across 10 runs. The shared initialization period (evaluations 1--20) is excluded to highlight differences during guided optimization. GA shows initial descent to $\approx3.4$ by evaluation 40, then continues gradual improvement to plateau near $3.1-3.2$ by evaluation 50--60, with minimal improvement afterward. All BOMs substantially outperform GA. Standard BO improves steadily but remains limited, reaching a median GEH of $1.28$ (Adaptive) and $1.37$ (Thompson) by evaluation 100.

TuRBO methods converge faster and finish at substantially lower GEH values. Among all methods, TuRBO with Thompson Sampling shows the strongest performance, reaching a median GEH of 1.01 by evaluation 100 with the tightest variance, indicating consistency across runs. Multi-TuRBO with Thompson Sampling ranks second at median 1.05, followed by MG-TuRBO with Thompson at 1.06. The Adaptive variants perform slightly worse: TuRBO (Adaptive) reaches median 1.11, MG-TuRBO (Adaptive) reaches 1.13, and Multi-TuRBO (Adaptive) reaches 1.16. In this 14D problem, MG-TuRBO's many-guide strategy does not show clear advantage over simpler trust-region methods.
Adaptive acquisition gives faster early progress for BO and Multi-TuRBO, especially between evaluations 30 and 50. For TuRBO, however, Thompson Sampling achieves the best final performance. For MG-TuRBO, Adaptive gives a lower and more stable final median than Thompson, which shows higher variability across runs. The acquisition strategy  affects performance in a method-dependent manner. For all trust-region methods, Thompson Sampling outperforms Adaptive: TuRBO improves by 9\%, Multi-TuRBO by 9\%, and MG-TuRBO by 6\%. This consistent pattern indicates that Thompson's aggressive exploitation pairs effectively with trust-region frameworks in 14D. Conversely, standard BO shows opposite behavior: Adaptive notably outperforms Thompson by 7\%, suggesting that global acquisition functions benefit from Adaptive's balanced exploration-exploitation trade-off.

\begin{figure}[!t]
\centering
\includegraphics[width=\columnwidth]{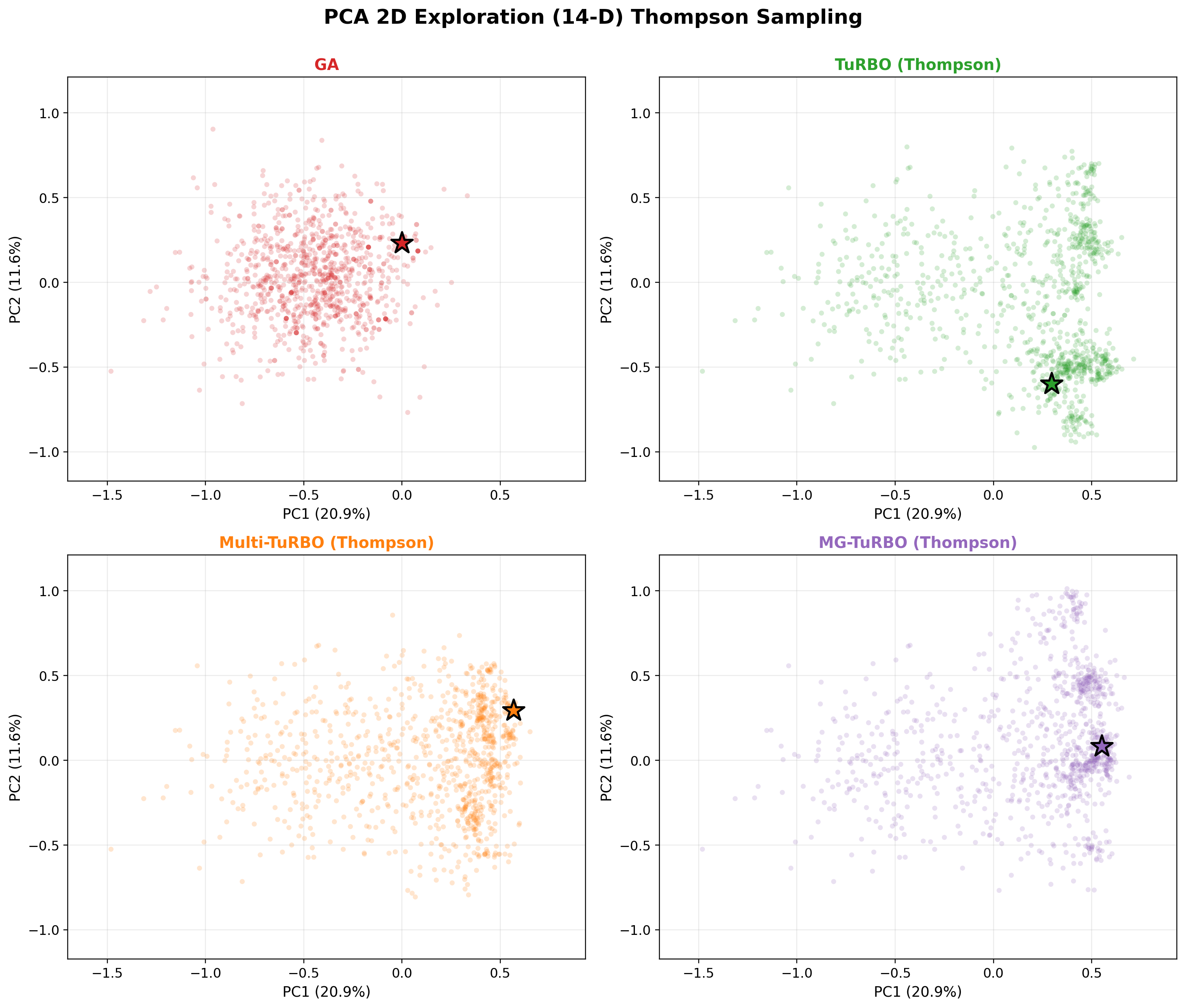}
\caption{PCA-based 2D exploration patterns for the 14D network comparing GA with TuRBO methods under Thompson Sampling. The best for each method is marked with a star.}
\label{fig:pca_14d}
\end{figure}

\begin{figure}[t]
\centering
\includegraphics[width=\columnwidth]{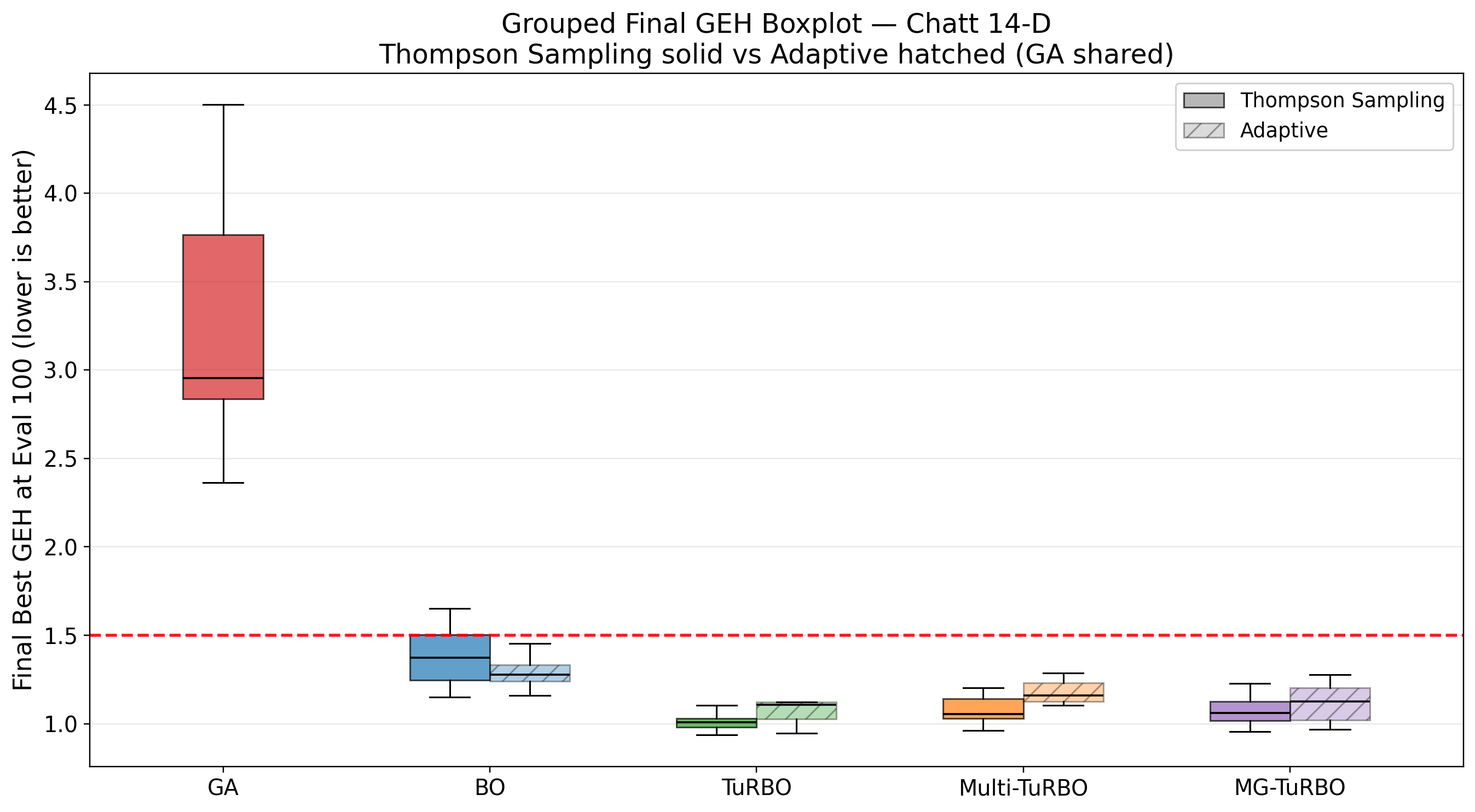}
\caption{Final best GEH distributions for Thompson Sampling and Adaptive acquisition across BOMs, with GA as a shared baseline.}
\label{fig:boxplot_14d}
\end{figure}

\cref{fig:pca_14d} shows PCA projections of the evaluated points under Thompson Sampling. The first two principal components explain 32.5\% of the total variance (PC1: 20.9\%, PC2: 11.6\%). All evaluated points from 10 runs are projected into a shared PCA space.  \cref{fig:pca_14d} and \cref{fig:convergence_14d} have consistent results. GA spreads evaluations broadly across the feasible region but shows little concentration around high-quality solutions. Standard BO forms a few main clusters, indicating partial concentration but weaker local refinement. TuRBO produces tight, dense clusters, consistent with focused local search within trust regions. Multi-TuRBO forms several distinct clusters, reflecting parallel exploration across multiple trust regions. MG-TuRBO shows similarly structured clustering, with more visible movement between regions. In the 14D case, this added structure does not translate into a clear performance gain over TuRBO. \cref{fig:boxplot_14d} summarizes the final best GEH at evaluation 100 for all methods. 

\cref{fig:boxplot_14d} confirms that TuRBO with Thompson Sampling is the strongest 14D method with with median GEH of 1.01 and lowest variance among all methods.  Multi-TuRBO with Thompson Sampling ranks second at median 1.06, followed by MG-TuRBO with Thompson at 1.08. For MG-TuRBO, Thompson also outperforms Adaptive by 4\% with comparable consistency. Adaptive variants show consistently higher medians and wider spreads. 
BO has the highest medians among BOMs at 1.29 (Adaptive) and 1.38 (Thompson), with wider variability but all runs remain below 1.7. 

\subsection{84 Dimensional Calibration Optimization}
\begin{figure}[t]
    \centering
    \includegraphics[width=\columnwidth]{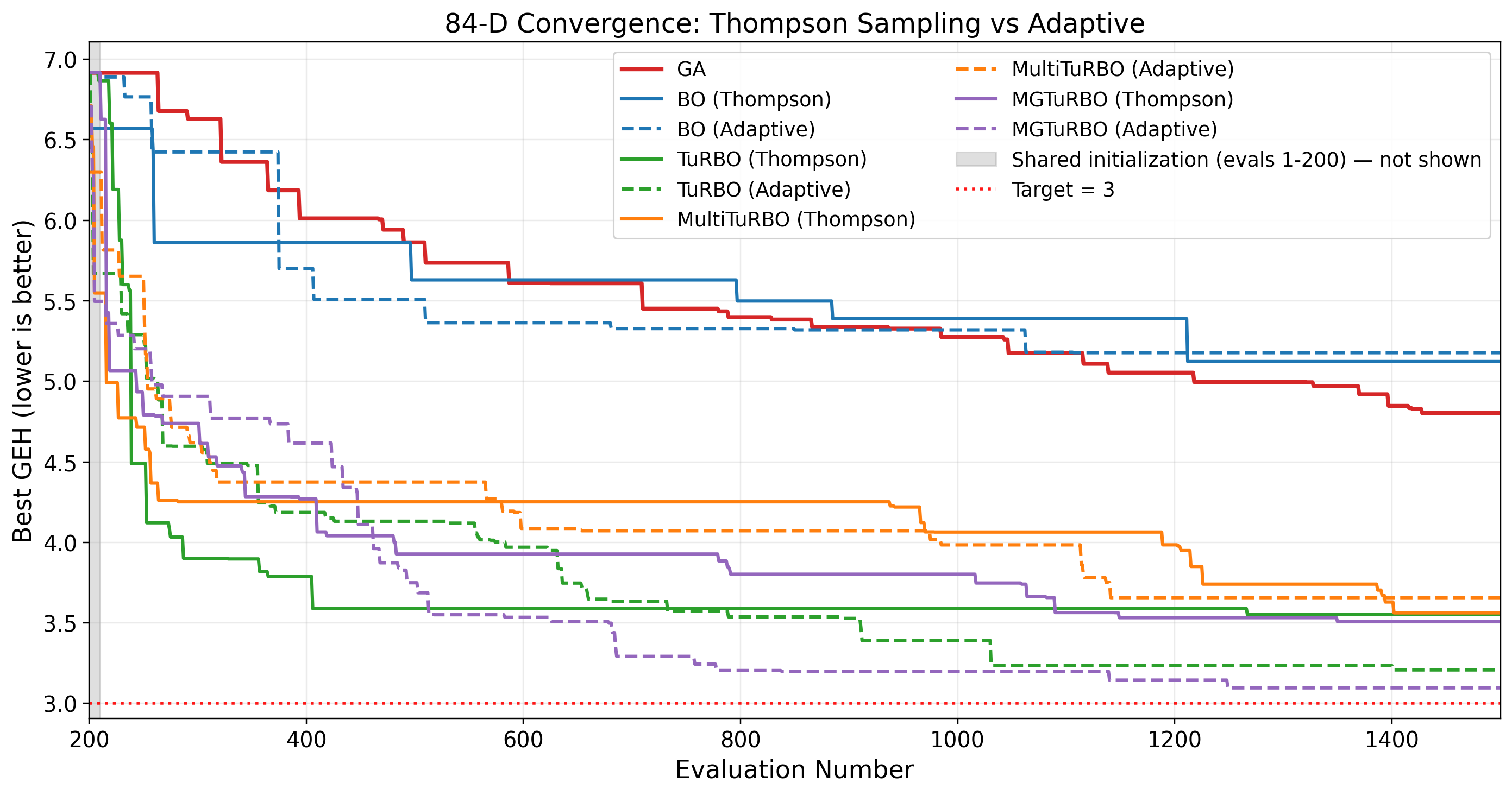}
    \caption{Optimization-phase convergence for the 84D calibration problem.} 
    \label{fig:convergence_84d}
\end{figure}

The 84D Nashville corridor is a much harder calibration problem than the 14D Chattanooga case. 
Due to the larger search space, BOMs use 200 initial samples before guided optimization begins. \cref{fig:convergence_84d} compares methods with Thompson Sampling and Adaptive acquisition during the optimization phase (evaluations 201+) for a representative single run. In this higher-dimensional setting, the relative ranking of methods changes clearly. MG-TuRBO with Adaptive acquisition performs best, reaching a final GEH of approximately 3.1 by evaluation 1500.

TuRBO with Adaptive acquisition ranks second at approximately 3.2, while TuRBO with Thompson reaches 3.6. Multi-TuRBO performs moderately under both acquisition strategies, reaching approximately 3.6--3.7 by evaluation 1500. These trust-region methods all substantially outperform both standard BO and GA. Standard BO makes limited progress after initialization for both acquisition strategies. GA converges to approximately 4.8, performing better than standard BO in this high-dimensional single-run comparison---though both remain far from the trust-region methods. The dramatic performance gap between trust-region methods and non-trust-region approaches underscores the critical importance of localized search in 84D. MG-TuRBO shows the clearest benefit from Adaptive acquisition (3.1 vs.\ 3.5 for Thompson), while TuRBO also favors Adaptive (3.2 vs.\ 3.6). Other methods show little sensitivity to acquisition-strategy. This pattern suggests that Adaptive acquisition particularly benefits methods employing aggressive multi-region exploration strategies in high dimensions.

\begin{figure}[t]
    \centering
    \includegraphics[width=\columnwidth]{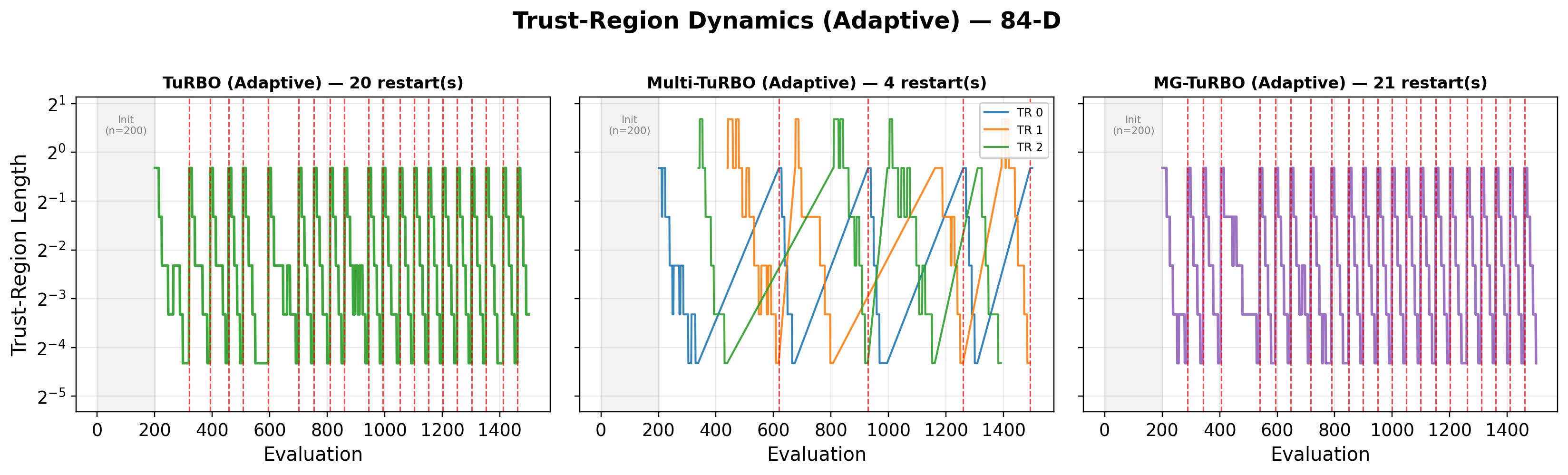}
    \caption{Trust-region search mechanisms in 84D (Adaptive acquisition). TuRBO executes 20 sequential restarts, each time exploring a single region until failure. Multi-TuRBO maintains concurrent regions exploring different basins. MG-TuRBO cycles through 21 local regions, allocating small budgets to systematically sample the landscape.} 
    \label{fig:trust_region_dynamics_84d}
\end{figure}

\cref{fig:trust_region_dynamics_84d} explains these performance differences by visualizing trust-region behavior for the representative 84D run using Adaptive acquisition across all three trust-region methods. TuRBO has 20 restarts, occurring after the trust region contracts without sufficient improvement---indicating persistent trapping in local basins. With only 1 active trust region, TuRBO spends many evaluations on local refinement before restarting. This proves inefficient in a vast 84D search space. Multi-TuRBO significantly reduces restart frequency to 4 major restart events by maintaining multiple trust regions in parallel. The active regions (shown in distinct colors) explore different parts of the space simultaneously, improving search efficiency compared to TuRBO's sequential approach. However, Multi-TuRBO can still allocate substantial budget refining moderately promising regions if initial region placements are suboptimal. MG-TuRBO has the most restarts (21). These do not indicate failure, but serve a fundamental and strategic role. MG-TuRBO intentionally allocates smaller budgets to each local region, proactively moving on after extracting sufficient gradient information. It then uses the updated global surrogate to select the next promising region for local refinement. This systematic rapid cycling allows MG-TuRBO to sample more basins across the 84D space instead of over-committing to any single region, ultimately achieving the best performance. 

Overall, the 84D results differ fundamentally from the 14D case. In 14D, focused single-region search with TuRBO (Thompson) performs best. In 84D, broader multi-basin exploration with MG-TuRBO (Adaptive) achieves superior performance, demonstrating that dimensional scaling fundamentally alters optimal search strategies. 
\section{Conclusion}

We compare optimization methods for traffic simulation calibration on two real-world problems: a 14D Chattanooga network and an 84D Nashville network. The results show that performance depends on problem dimension. In 14D, TuRBO with Thompson Sampling performed best, giving the lowest and most consistent GEH values. In this lower-dimensional setting, a focused single trust region was sufficient, and more complex multi-region strategies offered limited benefit. In 84D, MG-TuRBO with Adaptive acquisition performed best, with Multi-TuRBO also showing strong performance. In this higher-D setting, broader exploration across multiple regions became more important, while single-region TuRBO, standard BO, and GA were less effective under the same budget. Overall, the results suggest that lower-dimensional problems are handled well by simpler single-region methods such as TuRBO, while higher-dimensional problems benefit from broader multi-region search strategies, with MG-TuRBO showing the strongest performance.

\section{Limitations and Future Work}
This study considered only two problem sizes, 14D and 84D, so the transition between low- and high-dimensional settings remains unclear. In addition, the analysis focused on GEH as the calibration objective and on two networks with similar traffic characteristics. Future work should examine intermediate-dimensional problems, alternative calibration objectives, and a broader range of network types. It should also explore adaptive acquisition strategies, multi-fidelity calibration, and online calibration for real-time applications.
\section*{ACKNOWLEDGMENTS}
 We thank the City of Chattanooga and City of Nashville for sharing traffic demand and signal data.


\bibliographystyle{IEEEtran}
\bibliography{refs}

\end{document}